\documentclass[conference,a4paper]{APSIPA2026}

\usepackage[table]{xcolor}            
\newcommand{\best}[1]{\cellcolor{red!60}#1}
\newcommand{\second}[1]{\cellcolor{orange!57}#1}
\newcommand{\third}[1]{\cellcolor{yellow!55}#1}

\usepackage{hyperref}
\hypersetup{
    colorlinks=true,
    linkcolor=black,
    urlcolor=black,
    citecolor=black
}

\usepackage{amsmath,amssymb}
\usepackage{graphicx}
\usepackage{multirow}
\usepackage{threeparttable}
\usepackage{booktabs}
\usepackage{xspace}
\usepackage[backend=biber,style=ieee]{biblatex}
\usepackage{geometry}
\usepackage{fancyhdr}

\fancypagestyle{firststyle}{
  \fancyhf{}
  \fancyhead[C]{2026 Asia Pacific Signal and Information Processing Association Annual Summit and Conference (APSIPA ASC)}
}

\newcommand{\dgs}{D-3DGS\xspace}
\newcommand{\dnerf}{D-NeRF\xspace}
\newcommand{\nerfds}{NeRF-DS\xspace}
\newcommand{\cfc}{CfC\xspace}
\newcommand{\ltc}{LTC\xspace}

\begin{document}

\title{Stochastic Liquid Deformation Fields: An SDE Generalisation of Closed-form Continuous-time Cells for Dynamic 3D Gaussian Splatting}

\author{
\authorblockN{
Mingzhao Li\authorrefmark{1},\quad
Arghya Pal\authorrefmark{1}
}

\authorblockA{
\authorrefmark{1}
School of Information Technology, Monash University, Selangor, Malaysia\\
E-mail: \{mlii0259@student, arghya.pal\}@monash.edu}
}

\maketitle
\thispagestyle{firststyle}
\pagestyle{empty}

\begin{abstract}
Deformable 3D Gaussian Splatting (\dgs) reconstructs dynamic scenes by deforming a canonical set of 3D Gaussians through a deformation field of frame time. Replacing its MLP with a stack of Closed-form Continuous-time (\cfc) cells---a Liquid Neural Network that solves the Liquid Time-constant ODE in closed form---gives the field continuous-time behaviour at feed-forward cost. That closed form, however, is only the deterministic limit of a noise-driven system, and drops the stochastic term usually credited for the robustness of liquid networks. We put it back: a small Gaussian perturbation is added to the time gate of every \cfc cell, turning the deterministic field into a simple stochastic (SDE) one. The noise is used only during training, needs no solver, adds no parameters, and reduces exactly to the \cfc when switched off---so it costs nothing at inference. On the synthetic \dnerf scenes the stochastic field is on par with the deterministic \cfc and ahead of the MLP on most scenes; on the real-world \nerfds scenes the deterministic limit is already best and adding noise does not help. Our contribution is therefore a clean way to read the \cfc field as an SDE and an honest account of when a plain noise term helps and when it does not, rather than a new state of the art.
\end{abstract}


\section{Introduction}
\label{sec:intro}
Reconstructing a dynamic 3D scene from a single moving camera---4D reconstruction---is a foundational problem in vision and graphics. Explicit 3D Gaussian Splatting~\cite{kerbl3Dgaussians} achieves real-time photorealistic rendering, and Deformable 3DGS (\dgs)~\cite{yang2023deformable3dgs} lifts it to dynamic scenes by learning canonical Gaussians together with a positional-encoded MLP deformation field $F_\theta(\gamma(x),\gamma(t))\!\to\!(\Delta x,\Delta r,\Delta s)$.

\paragraph*{From discrete MLP to liquid field}
Although $t$ is continuous, the \dgs MLP couples no two timesteps in its architecture; it is fitted independently at each sampled frame, so temporal smoothness is left to optimisation. A natural fix replaces $F_\theta$ with a stack of Closed-form Continuous-time (\cfc) cells~\cite{hasani2022cfc}, the analytic solution of the Liquid Time-constant (\ltc) ODE~\cite{hasani2020ltn}. Each \cfc cell exposes a sigmoidal time gate that bakes a smooth, learned response to $t$ into the loss landscape, giving the field explicit continuous-time semantics while keeping inference a feed-forward pass---no numerical solver, unlike Neural-ODE methods such as ODE-GS~\cite{wang2025odegs}.

\paragraph*{The discarded stochasticity}
The \cfc is, by construction, the \emph{deterministic} limit of a noise-driven dynamical system: the \ltc ODE from which it derives was formulated as a stochastic process, and the robustness of liquid networks to jitter and irregular sampling~\cite{kumar2023lnn,karn2024lnnframework} is attributed precisely to that stochastic lineage. The closed form keeps the drift and drops the diffusion. Stochastic differential equation (SDE) models~\cite{li2020sde} retain the diffusion explicitly and are reported to be more robust under corrupted supervision, but only by paying a stochastic-solver cost on every forward pass. We seek the middle point: re-introduce the diffusion term \emph{architecturally}, inside the closed-form cell, without any solver.

\paragraph*{This paper}
We propose a \emph{stochastic liquid deformation field}. The only path through which the elapsed-time signal $\tau$ enters a \cfc cell is the affine pre-activation of its sigmoidal gate; we treat that pre-activation as an It\^{o} process and add a small Gaussian increment $\lambda\varepsilon$ ($\varepsilon\sim\mathcal{N}(0,\mathbf{I})$, with a single noise level $\lambda$) before the sigmoid---the resulting stochastic gate is~\eqref{eq:sdegate}. This is one Euler--Maruyama step of the SDE whose drift is the original deterministic gate, so $\lambda{=}0$ recovers the closed-form \cfc exactly, and the noise---injected only during training---leaves inference identical to the deterministic feed-forward pass. The numerical gains are modest; we present the method as a principled interpretation and an empirical characterisation rather than a new benchmark result.

\paragraph*{Contributions}
\begin{itemize}\setlength\itemsep{1pt}
\item A stochastic (SDE) reading of the closed-form \cfc deformation field: one Gaussian perturbation of the cell time gate~\eqref{eq:sdegate} restores the diffusion term the closed form drops. It needs no solver, adds no parameters, runs only during training, and reduces exactly to the \cfc at $\lambda{=}0$---so it is free at inference and the SDE-vs-\cfc comparison is exactly size-matched.
\item Results on \dnerf and \nerfds: the liquid field, with or without noise, is competitive with or ahead of the \dgs MLP on the synthetic scenes and matches it on the real-world scenes, together with a compute comparison (params, inference cost).
\item An honest, calibrated noise study: a small constant noise is harmless on synthetic data but does not help---and slightly hurts---on real-world data, so on these interpolation benchmarks the deterministic \cfc limit is already hard to beat.
\end{itemize}

\section{Related Work}
\label{sec:related}

\paragraph*{Dynamic 3D Gaussian splatting}
Following 3DGS~\cite{kerbl3Dgaussians}, \dgs~\cite{yang2023deformable3dgs} learns a canonical Gaussian set with an MLP deformation field; Gaussian-Flow~\cite{lin2024gaussianflow}, Shape of Motion~\cite{wang2025shapeofmotion} and FLAG-4D~\cite{tan2026flag4d} explore explicit flow and dual-deformation variants. None treats the deformation as an explicit \emph{stochastic} continuous-time function of $t$.

\paragraph*{Continuous-time and stochastic neural networks}
Neural ODEs~\cite{chen2018neuralode} parameterise a hidden-state derivative recovered by a solver; ODE-GS~\cite{wang2025odegs} applies a latent neural ODE to Gaussian-splat extrapolation, paying the solver cost on every forward pass. Neural SDEs~\cite{li2020sde} replace the deterministic dynamics with an It\^{o} integral, gaining robustness to observation noise at a still higher cost. Liquid Time-constant networks~\cite{hasani2020ltn} combine explicit time constants with input-dependent forcing, and \cfc~\cite{hasani2022cfc} derives the analytic, deterministic solution of that ODE, replacing the solver with a sigmoidal time gate. Liquid-S4~\cite{hasani2023lstate} brings liquid time constants to state-space models, while subsequent work characterises the robustness of liquid networks for noisy, irregular signals~\cite{kumar2023lnn,karn2024lnnframework}. Our method sits between the deterministic \cfc and a full Neural SDE: it restores the SDE's diffusion term inside the closed-form cell, keeping feed-forward inference.

\section{Method}
\label{sec:method}

\subsection{Background: Liquid Deformation Field}
\label{sec:method:bg}
A scene is $N$ canonical Gaussians $\{\mathcal{G}_i=(x_i,r_i,s_i,\alpha_i,c_i)\}$. For frame time $t\!\in\![0,1]$ a deformation field produces per-Gaussian offsets $(\Delta x_i,\Delta r_i,\Delta s_i)=F_\theta(\gamma(\mathrm{sg}(x_i)),\gamma(t))$, where $\gamma(\cdot)$ is a NeRF positional encoding and $\mathrm{sg}(\cdot)$ a stop-gradient. The liquid field instantiates $F_\theta$ as a stack of $D$ \cfc cells. A single cell maps an input $u$, hidden state $h$ and elapsed-time signal $\tau$ to an updated state through a shared backbone $\phi$, four linear heads, and a sigmoidal time gate:
\begin{equation}
\begin{aligned}
z &= \phi([u;\,h]),\\
g &= \tanh(W_g z),\quad h_{\text{cand}} = \tanh(W_h z),\\
\sigma_\tau &= \sigma\!\big(W_a z\,\tau + W_b z\big),\\
h' &= g\odot(1-\sigma_\tau)\;+\;h_{\text{cand}}\odot\sigma_\tau .
\end{aligned}
\label{eq:cfc}
\end{equation}
Equation~\eqref{eq:cfc} is the closed-form solution of the \ltc ODE~\cite{hasani2020ltn,hasani2022cfc}. The hidden state is reset to $\mathbf{0}$ each pass and threaded through the stack, with $t$ supplied as the elapsed-time signal $\tau$ of every cell; a linear head reads out the offsets. The canonical Gaussian set, rasteriser, $\mathcal{L}_1$+SSIM loss, density control and Adam schedule are inherited unchanged from \dgs~\cite{yang2023deformable3dgs}. Our contribution lives entirely inside the cell~\eqref{eq:cfc}.

\begin{figure*}[t]
\centering
\includegraphics[width=0.98\textwidth]{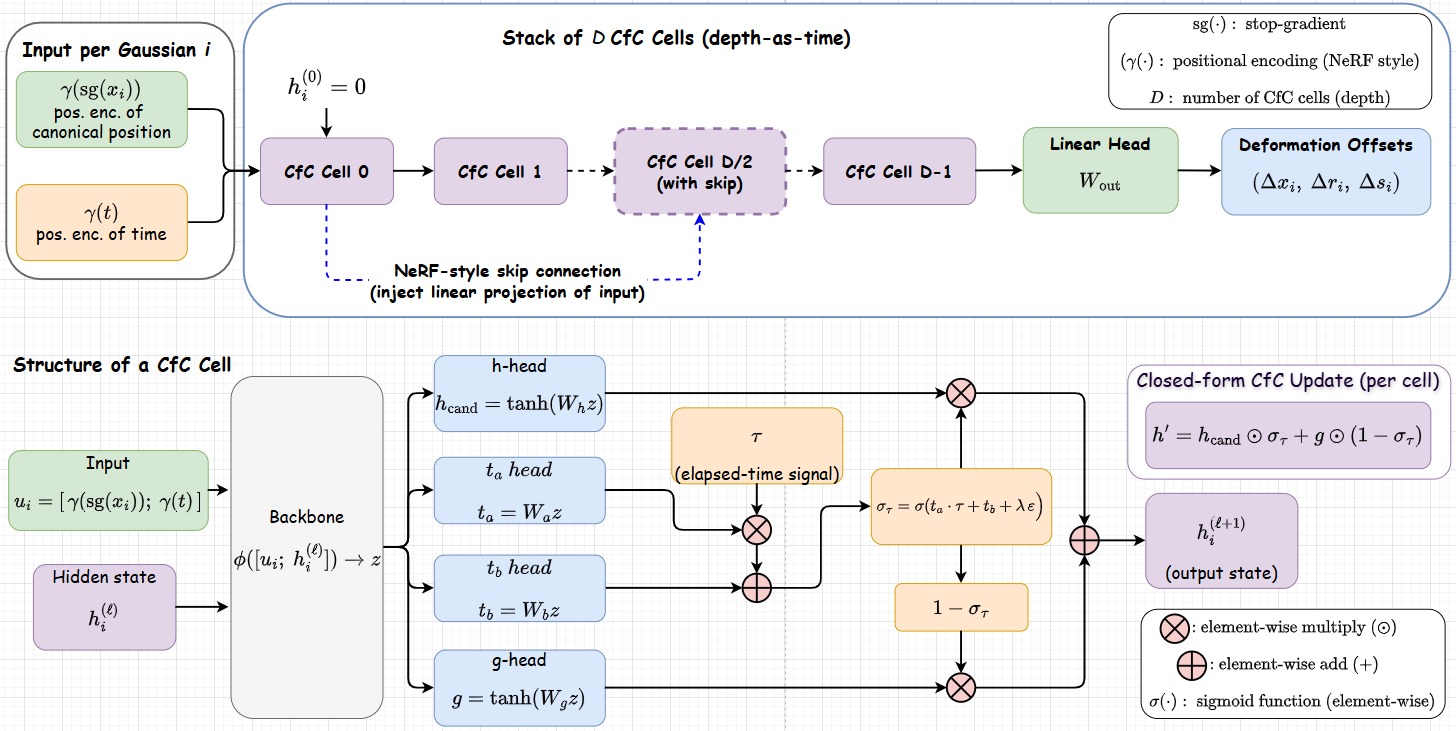}
\caption{\textbf{Stochastic liquid cell.} A shared backbone $\phi([u;h])\!\to\!z$ feeds four heads: $g,h_{\mathrm{cand}}$ form two candidate states, while $W_a z, W_b z$ form the affine elapsed-time term. The time gate is the only path through which $\tau$ enters the cell; we add a Gaussian diffusion increment $\lambda\varepsilon$ to its pre-activation (\,$\sigma_\tau=\sigma(W_a z\,\tau+W_b z+\lambda\varepsilon)$\,), an Euler--Maruyama step of the SDE whose drift is the deterministic gate. Noise is active in training only.}
\label{fig:cell}
\end{figure*}

\subsection{Stochastic (SDE) Time Gate}
\label{sec:method:sde}
The \ltc ODE underlying~\eqref{eq:cfc} is a noise-driven system; the closed form is its deterministic limit and therefore discards the diffusion. We restore it at the single point through which $\tau$ enters the cell. Writing the gate pre-activation as a latent state $s=W_a z\,\tau+W_b z$, we model it as a one-step It\^{o} process
\begin{equation}
\mathrm{d}s = \underbrace{(W_a z\,\tau + W_b z)}_{\text{drift}}\;+\;\underbrace{\lambda\,\mathrm{d}W}_{\text{diffusion}},
\label{eq:ito}
\end{equation}
whose Euler--Maruyama discretisation adds a Gaussian increment $\lambda\varepsilon$, $\varepsilon\sim\mathcal{N}(0,\mathbf{I})$ (the constant $\lambda$ absorbs $\sqrt{\Delta t}$). The stochastic cell is therefore~\eqref{eq:cfc} with the gate replaced by
\begin{equation}
\sigma_\tau=\sigma\!\big(W_a z\,\tau+W_b z+\lambda\,\varepsilon\big),\qquad \varepsilon\sim\mathcal{N}(0,\mathbf{I}).
\label{eq:sdegate}
\end{equation}
Three properties follow. \emph{(i) Exact reduction.} At $\lambda{=}0$ the increment vanishes and~\eqref{eq:sdegate} is identical to~\eqref{eq:cfc}; the stochastic field is a strict generalisation of the deterministic \cfc, and of the \dgs MLP through it. \emph{(ii) Solver-free, deterministic inference.} The noise is sampled per forward pass during training only; at test time it is disabled, so inference is the same feed-forward pass as the deterministic \cfc---no stochastic solver, unlike Neural SDEs~\cite{li2020sde}. \emph{(iii) Regularisation on $t$.} Because the gate is the only route by which $\tau$ acts, perturbing it perturbs exactly the field's response to time. In expectation the gate $\mathbb{E}_\varepsilon[\sigma_\tau]$ becomes a smoother function of $\tau$, so~\eqref{eq:sdegate} is \emph{intended} to act as a learnable smoothness prior---the mechanism credited for liquid-network robustness~\cite{kumar2023lnn}---with a strength set by $\lambda$. Whether this actually helps is an empirical question, which Sec.~\ref{sec:exp} answers.

\paragraph*{From the \ltc-SDE to one gate increment}
Our formulation is deliberately a \emph{first-order, single-step} realisation of the underlying stochastic dynamics, and we make the approximation chain explicit. The full liquid system is an SDE in the hidden state $h(t)$; its closed-form drift is exactly the deterministic \cfc update~\eqref{eq:cfc}. A faithful stochastic version would integrate a hidden-state diffusion over the elapsed interval with a solver---precisely the cost we set out to avoid. We instead inject a single Euler--Maruyama increment and place it on the gate pre-activation, because that scalar is the \emph{only} channel through which elapsed time $\tau$ enters the cell~\eqref{eq:cfc}: perturbing it is the minimal way to make the cell's time response stochastic without a solver and without touching the two candidate states $g,h_{\mathrm{cand}}$. In code this is one line---add $\lambda\varepsilon$ to $W_a z\,\tau+W_b z$ before the sigmoid, drawn independently per cell and per forward pass---which is exactly~\eqref{eq:sdegate}. We do not claim a higher-order or hidden-state SDE; those are the natural stronger variants, discussed in Sec.~\ref{sec:discussion}.

\subsection{Implementation}
\label{sec:method:impl}
We use $D{=}6$ cells, hidden width $W{=}128$, backbone depth $2$ and GELU activation, with a time read-out width of $32$ (\dnerf) / $64$ (\nerfds); the backbone width matches the deterministic liquid field. Only $\lambda$ is new. The noise is drawn independently per cell and per forward pass and disabled at inference, so it adds no parameters and no test-time cost (Sec.~\ref{sec:cost}); its training overhead is one Gaussian sample and one addition per cell. All other components follow the public \dgs pipeline (Sec.~\ref{sec:method:bg}). \dnerf scenes train for $40$k iterations and \nerfds for $20$k on a single GPU.

\section{Experiments}
\label{sec:exp}

\subsection{Setup}
We evaluate on \dnerf~\cite{pumarola2020dnerf} (eight synthetic monocular scenes at $800\!\times\!800$) and \nerfds~\cite{yan2023nerfds} (seven real-world scenes with specular, dynamic objects). We report PSNR, SSIM and LPIPS. Two baselines are retrained under an identical protocol and identical seeds: the original \dgs MLP field~\cite{yang2023deformable3dgs}, and the deterministic Liquid \cfc field, i.e.\ our model at $\lambda{=}0$. The proposed SDE-LNN uses the calibrated noise level $\lambda{=}0.05$ unless stated.

\paragraph*{Statistical calibration}
To gauge significance we retrained the identical $\lambda{=}0$ Hell~Warrior model three times; its PSNR spans $41.48$--$41.57$~dB ($\sigma\!\approx\!0.05$~dB). We therefore read aggregate differences below $\sim\!0.1$~dB as ties, and rely on (i) per-scene effects well above this floor (e.g.\ Hook, $+1.24$~dB) and (ii) the monotone $\lambda$ trend of Sec.~\ref{sec:abl} rather than single-run means.

\subsection{\dnerf: Eight Synthetic Scenes}
Table~\ref{tab:dnerf} reports per-scene results. Against the matched \dgs MLP baseline, the stochastic liquid field ($\lambda{=}0.05$) attains the best PSNR on six of the eight scenes and a higher mean ($38.28$ vs.\ $38.19$~dB), though the aggregate gap is within the variance above. Its clearest gains fall on the scenes with the most high-frequency articulated motion---Hook ($+1.24$~dB over the MLP), Hell~Warrior ($+0.38$), Stand~Up ($+0.37$) and Jumping~Jacks ($+0.15$)---consistent with the time-jitter regularisation of the stochastic gate~\eqref{eq:sdegate}; the MLP retains a small edge on Bouncing~Balls and T-Rex. This comparison mixes architecture with capacity (Table~\ref{tab:cost}); the effect of the noise alone is isolated in Sec.~\ref{sec:abl}. Fig.~\ref{fig:qual} shows the visible difference on Hell~Warrior, Hook and Stand~Up. All learned-deformation methods sit far above the implicit \dnerf and TiNeuVox baselines.

\begin{table*}[!t]
\centering
\caption{Per-scene comparison on the eight \dnerf synthetic scenes. PSNR$\,(\uparrow)$, SSIM$\,(\uparrow)$ and LPIPS$\,(\downarrow)$ at $800\!\times\!800$ test images. \dnerf and TiNeuVox numbers are quoted from~\cite{yang2023deformable3dgs}; \dgs (MLP) and Ours (SDE-LNN, $\lambda{=}0.05$) are retrained under an identical protocol with the same canonical-Gaussian set and seeds. Cells are coloured \colorbox{red!60}{best}, \colorbox{orange!57}{second-best}, \colorbox{yellow!55}{third-best}.}
\label{tab:dnerf}
\setlength{\tabcolsep}{3pt}
\footnotesize
\begin{tabular}{l|ccc|ccc|ccc|ccc}
\toprule
& \multicolumn{3}{c|}{Hell Warrior} & \multicolumn{3}{c|}{Mutant} & \multicolumn{3}{c|}{Hook} & \multicolumn{3}{c}{Bouncing Balls}\\
Method & PSNR$\,\uparrow$ & SSIM$\,\uparrow$ & LPIPS$\,\downarrow$ & PSNR$\,\uparrow$ & SSIM$\,\uparrow$ & LPIPS$\,\downarrow$ & PSNR$\,\uparrow$ & SSIM$\,\uparrow$ & LPIPS$\,\downarrow$ & PSNR$\,\uparrow$ & SSIM$\,\uparrow$ & LPIPS$\,\downarrow$\\
\midrule
\dnerf~\cite{pumarola2020dnerf} & 24.06 & 0.9440 & \third{0.0707} & 30.31 & \third{0.9672} & \third{0.0392} & 29.02 & 0.9595 & \third{0.0546} & 38.17 & 0.9891 & \third{0.0323} \\
TiNeuVox~\cite{fang2022tineuvox} & \third{27.10} & \third{0.9638} & 0.0768 & \third{31.87} & 0.9607 & 0.0474 & \third{30.61} & \third{0.9599} & 0.0592 & \second{40.23} & \third{0.9926} & 0.0416 \\
\dgs (MLP)~\cite{yang2023deformable3dgs} & \second{41.23} & \second{0.9867} & \best{0.0255} & \second{42.05} & \best{0.9944} & \best{0.0067} & \second{36.94} & \second{0.9855} & \second{0.0167} & \best{41.37} & \best{0.9957} & \best{0.0086} \\
Ours (SDE-LNN) & \best{41.61} & \best{0.9873} & \second{0.0259} & \best{42.08} & \second{0.9941} & \second{0.0076} & \best{38.18} & \best{0.9884} & \best{0.0144} & \third{40.22} & \second{0.9950} & \second{0.0100} \\
\midrule
& \multicolumn{3}{c|}{Lego} & \multicolumn{3}{c|}{T-Rex} & \multicolumn{3}{c|}{Stand Up} & \multicolumn{3}{c}{Jumping Jacks}\\
Method & PSNR$\,\uparrow$ & SSIM$\,\uparrow$ & LPIPS$\,\downarrow$ & PSNR$\,\uparrow$ & SSIM$\,\uparrow$ & LPIPS$\,\downarrow$ & PSNR$\,\uparrow$ & SSIM$\,\uparrow$ & LPIPS$\,\downarrow$ & PSNR$\,\uparrow$ & SSIM$\,\uparrow$ & LPIPS$\,\downarrow$\\
\midrule
\dnerf~\cite{pumarola2020dnerf} & \second{25.56} & \third{0.9363} & \third{0.0821} & 30.61 & \third{0.9671} & 0.0535 & 33.13 & 0.9781 & 0.0355 & 32.70 & \third{0.9779} & \third{0.0388} \\
TiNeuVox~\cite{fang2022tineuvox} & \best{26.64} & 0.9258 & 0.0877 & \third{31.25} & 0.9666 & \third{0.0478} & \third{34.61} & \third{0.9797} & \third{0.0326} & \third{33.49} & 0.9771 & 0.0408 \\
\dgs (MLP)~\cite{yang2023deformable3dgs} & 24.92 & \second{0.9434} & \second{0.0445} & \best{37.73} & \best{0.9929} & \best{0.0103} & \second{43.85} & \second{0.9942} & \best{0.0083} & \second{37.40} & \second{0.9893} & \best{0.0141} \\
Ours (SDE-LNN) & \third{24.94} & \best{0.9436} & \best{0.0443} & \second{37.46} & \second{0.9926} & \second{0.0110} & \best{44.22} & \best{0.9944} & \second{0.0085} & \best{37.55} & \best{0.9896} & \second{0.0142} \\
\bottomrule
\end{tabular}
\end{table*}

\subsection{\nerfds: Seven Real-World Scenes}
Table~\ref{tab:nerfds} reports the seven real-world \nerfds scenes, which are dominated by pose noise so that the \emph{relative} ordering of methods is the more informative quantity. The stochastic field is competitive with the matched MLP in aggregate (mean PSNR $23.73$ vs.\ $23.82$~dB) and improves the hardest manipulation scenes, taking the best PSNR on Bell and Press ($+0.07$ and $+0.14$~dB over the MLP) and the best SSIM on three scenes (Plate, Bell, Press). It trails the MLP on the cleaner Cup and As scenes, which brings the aggregate to near-parity rather than a net gain---a behaviour we examine in Sec.~\ref{sec:abl}. Both liquid and MLP fields clearly surpass the specular-aware \nerfds and TiNeuVox baselines on mean PSNR.

\begin{table*}[!t]
\centering
\caption{Per-scene comparison on the seven \nerfds real-world scenes. NeRF-DS and TiNeuVox numbers are quoted from~\cite{yang2023deformable3dgs}; \dgs (MLP) and Ours (SDE-LNN, $\lambda{=}0.05$) are retrained under an identical protocol. Colouring as in Table~\ref{tab:dnerf}.}
\label{tab:nerfds}
\setlength{\tabcolsep}{3pt}
\footnotesize
\begin{tabular}{l|ccc|ccc|ccc|ccc}
\toprule
& \multicolumn{3}{c|}{Sieve} & \multicolumn{3}{c|}{Plate} & \multicolumn{3}{c|}{Bell} & \multicolumn{3}{c}{Press}\\
Method & PSNR$\,\uparrow$ & SSIM$\,\uparrow$ & LPIPS$\,\downarrow$ & PSNR$\,\uparrow$ & SSIM$\,\uparrow$ & LPIPS$\,\downarrow$ & PSNR$\,\uparrow$ & SSIM$\,\uparrow$ & LPIPS$\,\downarrow$ & PSNR$\,\uparrow$ & SSIM$\,\uparrow$ & LPIPS$\,\downarrow$\\
\midrule
NeRF-DS~\cite{yan2023nerfds} & \best{25.78} & \best{0.8900} & \best{0.1472} & \second{20.54} & \third{0.8042} & \best{0.1996} & \third{23.19} & 0.8212 & \third{0.1867} & \best{25.72} & \second{0.8618} & \third{0.2047} \\
TiNeuVox~\cite{fang2022tineuvox} & 21.49 & 0.8265 & 0.3176 & \best{20.58} & 0.8027 & 0.3317 & 23.08 & \third{0.8242} & 0.2568 & 24.47 & \third{0.8613} & 0.3001 \\
\dgs (MLP)~\cite{yang2023deformable3dgs} & \third{25.28} & \second{0.8721} & \second{0.1514} & \third{20.50} & \second{0.8140} & \second{0.2265} & \second{25.07} & \second{0.8389} & \best{0.1637} & \third{25.35} & \second{0.8618} & \best{0.1939} \\
Ours (SDE-LNN) & \second{25.43} & \third{0.8690} & \third{0.1627} & 20.43 & \best{0.8142} & \third{0.2314} & \best{25.14} & \best{0.8399} & \second{0.1702} & \second{25.49} & \best{0.8643} & \second{0.2015} \\
\midrule
& \multicolumn{3}{c|}{Cup} & \multicolumn{3}{c|}{As} & \multicolumn{3}{c|}{Basin} & \multicolumn{3}{c}{Mean}\\
Method & PSNR$\,\uparrow$ & SSIM$\,\uparrow$ & LPIPS$\,\downarrow$ & PSNR$\,\uparrow$ & SSIM$\,\uparrow$ & LPIPS$\,\downarrow$ & PSNR$\,\uparrow$ & SSIM$\,\uparrow$ & LPIPS$\,\downarrow$ & PSNR$\,\uparrow$ & SSIM$\,\uparrow$ & LPIPS$\,\downarrow$\\
\midrule
NeRF-DS~\cite{yan2023nerfds} & \best{24.91} & \third{0.8741} & \second{0.1737} & \third{25.13} & \third{0.8778} & \best{0.1741} & \second{19.96} & \best{0.8166} & \best{0.1855} & \third{23.60} & \second{0.8494} & \second{0.1816} \\
TiNeuVox~\cite{fang2022tineuvox} & 19.71 & 0.8109 & 0.3643 & 21.26 & 0.8289 & 0.3967 & \best{20.66} & \second{0.8145} & 0.2690 & 21.61 & 0.8241 & 0.3195 \\
\dgs (MLP)~\cite{yang2023deformable3dgs} & \second{24.81} & \best{0.8893} & \best{0.1560} & \best{26.19} & \best{0.8833} & \second{0.1814} & \third{19.56} & \third{0.7956} & \second{0.1882} & \best{23.82} & \best{0.8507} & \best{0.1802} \\
Ours (SDE-LNN) & \third{24.08} & \second{0.8793} & \third{0.1864} & \second{25.99} & \second{0.8810} & \third{0.1901} & 19.55 & 0.7902 & \third{0.2025} & \second{23.73} & \third{0.8483} & \third{0.1921} \\
\bottomrule
\end{tabular}
\end{table*}

\subsection{Compute and Overhead}
\label{sec:cost}
The stochastic term is free at inference. Because the noise is switched off at test time (Sec.~\ref{sec:method:sde}), SDE-LNN executes the \emph{identical} feed-forward pass as the deterministic \cfc, at the same throughput; it adds no parameters and, during training, only one Gaussian draw and one addition per cell. Table~\ref{tab:cost} lists the measured deformation-field sizes. Two points follow. First, the ``does noise help'' comparison (SDE-LNN vs.\ \cfc) is exactly size-matched, so the neutral result of Sec.~\ref{sec:abl} is not a capacity artifact. Second, the liquid field's own size is a design choice, not fixed by the method: on \nerfds it is $39\%$ smaller than the MLP, whereas the \dnerf runs use a wider backbone. All three fields keep feed-forward inference and need no numerical solver.

\begin{table}[!t]
\centering
\caption{Deformation-field parameter counts, measured with \texttt{ptflops}~\cite{flops2018ptflops} on the exact configurations used. The deterministic \cfc and SDE-LNN are architecturally identical---the training-time noise adds no parameters and is disabled at inference---so they share a row. The MLP count is the \dgs deformation field~\cite{yang2023deformable3dgs}.}
\label{tab:cost}
\setlength{\tabcolsep}{5pt}
\footnotesize
\begin{tabular}{l|c|c}
\toprule
Deformation field & Params (M) & Inference \\
\midrule
\dgs MLP ($D{=}8,W{=}256$)~\cite{yang2023deformable3dgs} & 0.52 & feed-forward \\
Liquid \cfc\,/\,SDE-LNN --- \nerfds & \best{0.32} & feed-forward \\
Liquid \cfc\,/\,SDE-LNN --- \dnerf & 1.55 & feed-forward \\
\bottomrule
\end{tabular}
\end{table}

\subsection{Noise-level Study}
\label{sec:abl}
Table~\ref{tab:abl} sweeps $\lambda$ and reports the dataset means, and the picture is simple: the noise never buys an aggregate gain. On \dnerf a small $\lambda$ is harmless---the mean is unchanged up to $\lambda{=}0.05$---while a larger $\lambda$ slowly washes out real motion and the mean falls. On \nerfds the deterministic limit ($\lambda{=}0$) already gives the best PSNR, and every positive $\lambda$ is a little worse. So a plain constant noise does not improve real-world reconstruction: the deterministic \cfc is the sweet spot, and $\lambda{=}0.05$ is just the most noise one can add on synthetic data before quality starts to fall.

\begin{table}[!t]
\centering
\caption{Noise-level study: dataset-mean metrics as a function of $\lambda$. $\lambda{=}0$ is the deterministic \cfc. \colorbox{red!60}{best} per column.}
\label{tab:abl}
\setlength{\tabcolsep}{4pt}
\footnotesize
\begin{tabular}{l|ccc|ccc}
\toprule
& \multicolumn{3}{c|}{\dnerf (8 scenes)} & \multicolumn{3}{c}{\nerfds (7 scenes)}\\
$\lambda$ & PSNR & SSIM & LPIPS & PSNR & SSIM & LPIPS\\
\midrule
0.00 & \best{38.28} & \best{0.9856} & \best{0.0168} & \best{23.83} & 0.8482 & 0.1908 \\
0.05 & \best{38.28} & \best{0.9856} & 0.0170 & 23.73 & 0.8483 & 0.1921 \\
0.10 & 38.06 & 0.9853 & 0.0173 & 23.73 & 0.8480 & 0.1912 \\
0.20 & 37.91 & 0.9854 & 0.0175 & 23.80 & \best{0.8489} & \best{0.1903} \\
\bottomrule
\end{tabular}
\end{table}

\subsection{Qualitative Results}
\begin{figure*}[t]
\centering
\makebox[0.273\linewidth]{\small Ground truth}\makebox[0.273\linewidth]{\small D-3DGS (MLP)}\makebox[0.273\linewidth]{\small Ours (SDE-LNN)}\\[2pt]
\includegraphics[width=0.82\linewidth,keepaspectratio]{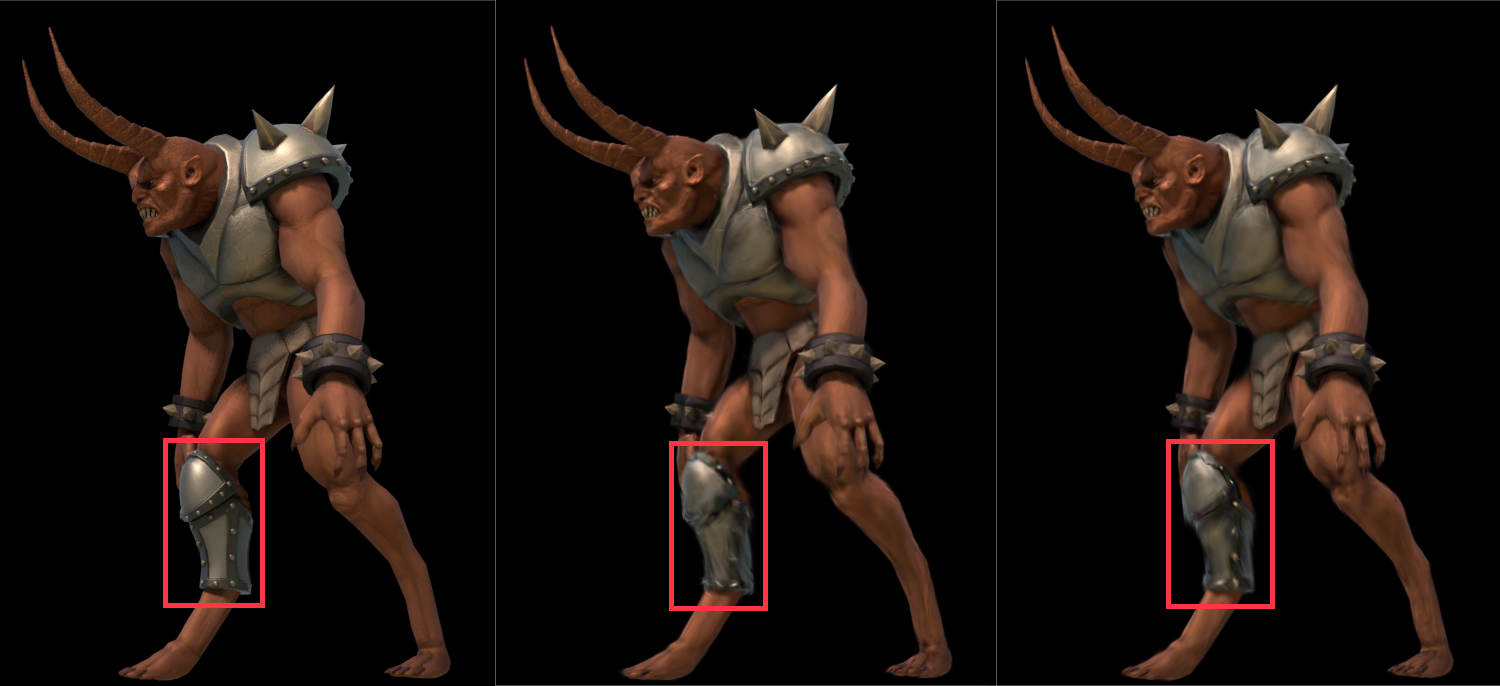}\\[1pt]
{\small (a) Hell Warrior ($+0.38$~dB): the MLP blurs the swung limb; SDE-LNN keeps its outline.}\\[5pt]
\includegraphics[width=0.82\linewidth,keepaspectratio]{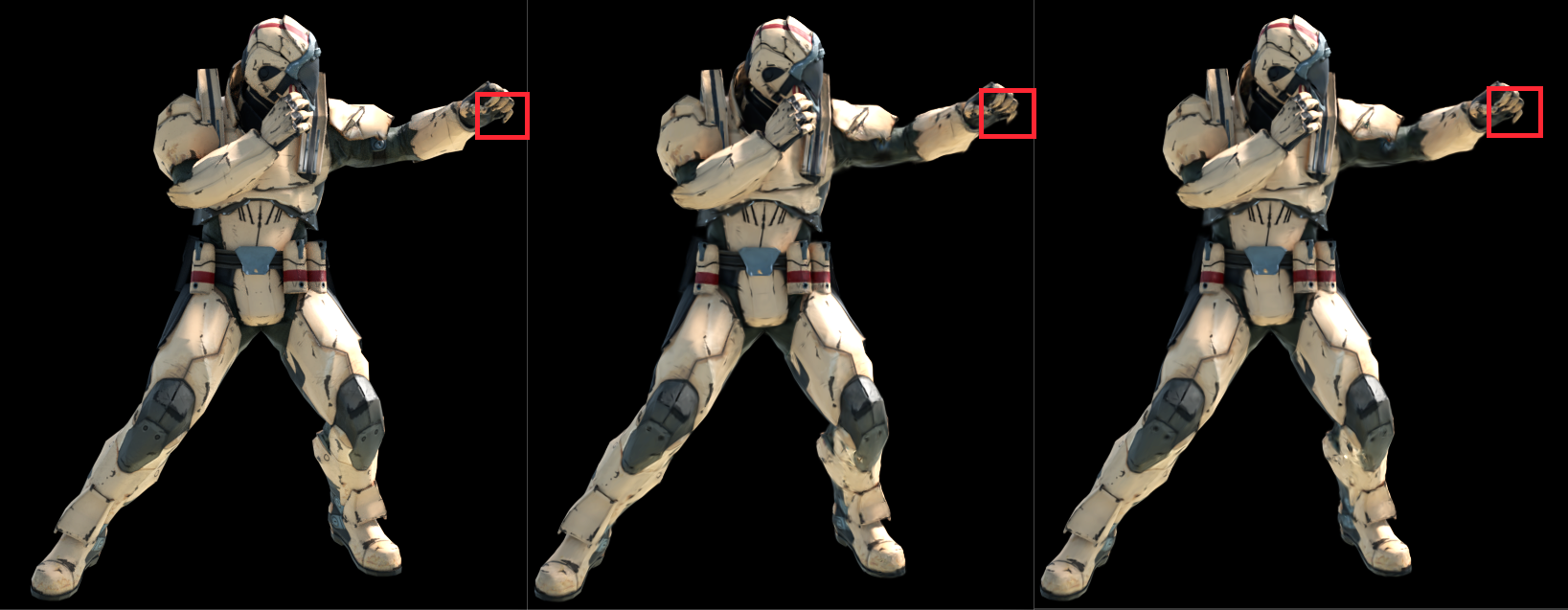}\\[1pt]
{\small (b) Hook ($+1.24$~dB, the largest gain in Table~\ref{tab:dnerf}): the fast arm is ghosted by the MLP, crisp with SDE-LNN.}\\[5pt]
\includegraphics[width=0.82\linewidth,keepaspectratio]{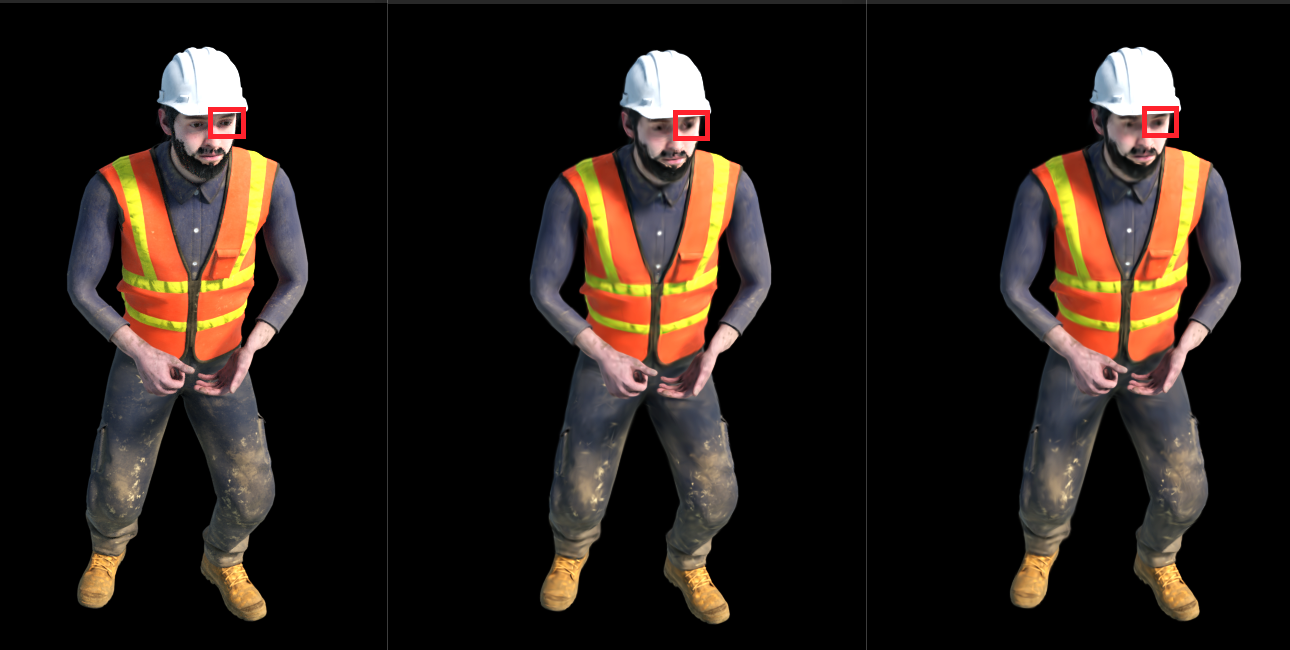}\\[1pt]
{\small (c) Stand Up ($+0.37$~dB): the eyebrow and facial detail are blurred by the MLP but sharper with SDE-LNN.}
\caption{\textbf{Qualitative comparison on \dnerf.} The three scenes with the largest PSNR gain over the MLP (Table~\ref{tab:dnerf}). Each row is one scene; within a row, left to right: \textbf{ground truth}, \textbf{\dgs (MLP)}, \textbf{Ours (SDE-LNN)}. Look at the fast-moving limb in each scene, where the MLP smears or ghosts the motion.}
\label{fig:qual}
\end{figure*}

Fig.~\ref{fig:qual} compares the three \dnerf scenes with the largest PSNR gain over the MLP. The differences are localised to the fast-moving parts: on each scene the MLP smears or leaves a faint trailing ghost around the articulated limb, while the stochastic liquid field keeps that edge sharper and closer to the ground-truth silhouette. This is the visible counterpart of the numbers in Table~\ref{tab:dnerf}.

\section{Discussion}
\label{sec:discussion}

\paragraph*{Why the noise does not help here}
These benchmarks supervise densely in time and ask only for interpolation \emph{inside} the observed window---a regime the deterministic \cfc already fits well, so a noise term has little error to correct, and on the noisier real-world scenes it does more harm than good. The plain constant noise is thus neutral at best: an honest negative result in which the SDE view's value is a clean formulation and a free, tunable knob, not an automatic quality gain.

\paragraph*{What would help: adaptive stochasticity}
Noise helps only where supervision is most jittery---fast \dnerf motion, a few hard \nerfds scenes---which is what it is meant to damp. A smarter noise than our constant one should do more: \emph{annealing} $\lambda$ to zero late in training, making $\lambda$ \emph{input-dependent}, or moving the diffusion from the gate to the \emph{hidden state} (a closer match to the true \ltc-SDE). All keep the solver-free, deterministic-inference property.

\paragraph*{Robustness regime and limitations}
The clean, dense \dnerf/\nerfds protocol under-tests the setting the noise targets; the informative stress test is corrupted or sparse supervision (jittered timestamps, pose noise, few-view training). We leave three additions to future work: this stress test; a qualitative comparison on the real-world scenes, where the small Bell and Press wins are best shown together; and inference-time Monte-Carlo sampling for uncertainty. We study only one constant, gate-level $\lambda$; on clean \dnerf the method gaps sit within the run-to-run variance of Sec.~\ref{sec:exp}.

\section{Conclusion}
\label{sec:conclusion}
We presented a stochastic view of the closed-form \cfc deformation field for Deformable 3D Gaussian Splatting: treating the cell time gate as a one-step SDE and adding a Gaussian noise to it restores the diffusion term the closed form drops, with no solver, no parameters, and deterministic inference. On \dnerf the resulting field matches the deterministic \cfc and is ahead of the MLP; on \nerfds the deterministic limit stays best and a plain constant noise does not help. The take-away is twofold: the \cfc field has a clean SDE interpretation with a free noise knob, and, on standard interpolation benchmarks, that knob is best left near zero. Making the noise pay off on noisy, real-world supervision---through annealed, input-dependent, or hidden-state schedules---is the natural next step.

\printbibliography

\end{document}